\documentclass[final,1p,times,twocolumn]{elsarticle}
\usepackage{epsfig}
\usepackage{amssymb, amsmath}
\usepackage{subfig}
\usepackage{multicol}
\usepackage{multirow}
\journal{Engineering Applications of Artificial Intelligence}

\begin{document}

\begin{frontmatter}



\title{Location Forensics of Media Recordings Utilizing Cascaded SVM and Pole-matching Classifiers}


\author{Jayanta~Dey, Mohammad~Ariful~Haque}

\address{deyjayanta76@gmail.com, arifulhoque@eee.buet.ac.bd}

\begin{abstract}
Information regarding the location of power distribution grid can be extracted from the power signature embedded in the multimedia signals (e.g., audio, video data) recorded near electrical activities. This implicit mechanism of identifying the origin-of-recording can be a very promising tool for multimedia forensics and security applications. In this work, we have developed a novel grid-of-origin identification system from media recording that consists of a number of support vector machine (SVM) followed by pole-matching (PM) classifiers. First, we determine the nominal frequency of the grid ($50$ or $60$ Hz) based on the spectral observation. Then an SVM classifier, trained for the detection of a grid with a particular nominal frequency, narrows down the list of possible grids on the basis of different discriminating features extracted from the electric network frequency (ENF) signal. The decision of the SVM classifier is then passed to the PM classifier that detects the final grid based on the minimum distance between the estimated poles of test and training grids. Thus, we start from the problem of classifying grids with different nominal frequencies and simplify the problem of classification in three stages based on nominal frequency, SVM and finally using PM classifier. This cascaded system of classification ensures better accuracy ($15.57$\% higher) compared to traditional ENF-based SVM classifiers described in the literature.
\end{abstract}


\end{frontmatter}


\section{Introduction}

With the proliferation of terrorism, child pornography \citep{unicef2011child} or abuse on women, location forensics has become an important area of research in the 21\textsuperscript{st} century. Success in identifying such locations properly can ease the process of getting hold of the criminals involved. Furthermore, correct estimation of recording location of media recordings may pave the way for automatic tagging of geo-location of huge amount of digital data which are being uploaded every moment on social media platforms such as YouTube and Facebook \citep{garg2013geo}. Therefore, a consistent as well as fast method of region-of-recording recognition is highly necessary.

A potential route to obtain the information of recording location from media recordings is to extract the fingerprint left on that recording by the power grid of that location. Due to electromagnetic radiation, every grid acts like an antenna that radiates an electromagnetic wave with a frequency similar to the corresponding grid frequency \citep{fechner2014humming}. It has a nominal value of 60 Hz in North America and 50 Hz in most other places of the world. This radiation introduces a weak interference in audio or video clips recorded near places where there is electrical activity. The instantaneous frequency of a power grid fluctuates over time around its nominal value due to control mechanism and load variation. The variation pattern of the Electric Network Frequency (ENF) over time for any grid is defined as an ENF signal. ENF can vary with time over a certain range for a particular grid. For example, in Lebanon, it varies by about 1 Hz around 50 Hz nominal frequency, whereas in China, this variation is about 0.1 Hz. The frequency variation pattern is almost identical in all places of the same grid \citep{fechner2014humming} and depends on load variation pattern of that grid. The load variation has different statistical distribution for different grids and follows a specific pattern for a particular grid. From the aforementioned facts, it can be inferred that the electromagnetic interference contains information about the grid. If such interference pattern can be reliably extracted from the recording, we can use this information to correctly identify the location of the recording. 

The ENF signal extracted from the media recording is generally utilized for grid-location identification. However, accurate and reliable estimation of ENF signal from the media recording is not a trivial task. The frequency of the electromagnetic interference signal needs to be estimated from short segments of media and sensor recordings that may contain human speech, acoustic noise, and other interfering signals. The amounts of noise and distortions can be different even within the same signal due to change of recording conditions which can heavily distort the actual ENF signal pattern. A variety of ENF extraction techniques have been reported in the literature. Initial work in this area was proposed by Grigoras and Cooper \citep{grigoras2005digital, cooper2008electric, grigoras2007applications, cooper2009automated, grigoras2009applications, cooper2011further}. Their study reveals that a reference ENF measurement obtained from anywhere in the grid can then be used to determine the time at which the recording was made and to detect tempering. Some of the recent works that have extended the initial research includes the extraction of ENF from video flicker \citep{garg2011seeing}, refining the Fourier approaches \citep{liu2012application}, using dynamic programming and a feed forward spectral estimator \citep{ojowu2012enf} and incorporating power signal harmonics \citep{hajj2013spectrum}. Most of the techniques are based on either time or frequency based methods or variations of these techniques. Both the approaches have practical applications. Time-based techniques such as zero-crossing method have been proven to be very useful in order to record the ENF directly from the power line. Such methods are used by power suppliers, which are obliged to keep the ENF within a given tolerance and thus need to record the ENF time history to validate that. Zero crossing method is, however, not suitable to extract ENF from real-world speech or music audio content. Spectrum or short-time-Fourier-transform (STFT) based methods, in contrast, are suitable for this, and are commonly applied for this purpose \citep{hajj2013spectrum}.  In short, there is no unique way to extract the ENF signal and different extraction methods may give different ENF estimations for the same media recording. 

A number of algorithms have been proposed in the literature to identify the grid location from the estimated ENF. Garg \textit{et al.} proposed a half-plane intersection method based on highly quantized information from the correlation coefficient between the processed ENF signals across different locations \citep{garg2013geo}. Unfortunately, the method provides satisfactory results only for power recordings since the audio data do not have high correlation. Most of the current techniques are machine learning based that rely on the extracted features from the ENF. However, ENF only captures the frequency variation of the underlying grid voltage. The power grid experiences different phenomena that not only changes the frequency but also the voltage of the grid. Therefore, complimentary information can be extracted from the raw data and utilizing these features along with the ENF-based classification pipeline may improve the grid identification accuracy. 

In this paper, we propose a novel grid-location identification method that not only uses the informations from the ENF signal but also the features obtained directly from the raw data. In our approach, SVM classifiers are trained on the feature extracted from the ENF signal of audio and power recordings with $50$ and $60$ Hz nominal frequency. In addition to that we estimate the AR parameters of short data segments of the training data set. The roots of the AR-parameter polynomial provide the poles of the AR model. In testing phase, the nominal frequency of the grid is identified based on spectral observation and then the particular SVM trained for the nominal frequency gives the most probable grids. Then poles are extracted from the testing data. Now a distance-based classifier is used to match the poles of the testing data with those of most probable grids of training data set. The grid with minimum average distance between the training and testing poles is estimated as the identified grid. We have tested our system using the data set provided for the IEEE Signal Processing Cup, 2016 \citep{spcup} on ``Location Forensic of Media Recording''. The results show that the proposed classification system can classify power/audio recordings with a higher accuracy of $95.08\%$ on the test set compared to that of $79.51$\% of only ENF-based SVM classifier.

The organization of rest of the paper is as follows. Section II defines the problem statement and provides a concise overview on the dataset. The proposed cascaded classification system is described in detail in Section III. Section IV shows the experimental results and comparison with a traditional ENF based clssification method. The paper is concluded with some remarks in Section V.

\begin{figure*}[h!]
\centering
\includegraphics[width = 3.5 in, height = 1.8 in]{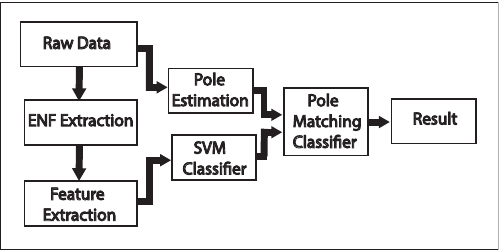}
\caption{Block diagram of the proposed classification system.}
\label{fig:layout}
\end{figure*}

\section{Data Description and Problem Statement}
In this work, the data provided for the IEEE Signal Processing Cup, 2016 \citep{spcup} have been used. Two types of signals namely audio and power signals are provided for training purpose from 9 different grids and the grids were labeled as A to I. Again for the grids labeled J to L only power recordings were provided. Table \ref{Grid_info} depicts the location of the grids of the dataset along with their corresponding labels. The data consist of 8 grids with nominal frequency of 50 Hz and 4 grids with nominal frequency of 60 Hz. Power recordings were taken by a signal recorder connected to a power outlet	 using a step-down transformer. However, the audio signal is simply recorded using an audio recorder not connected to anything.  
\begin{table*}[t]
	\centering
	\caption{Location of the grids in the dataset.}\label{Grid_info}
	\begin{tabular}{|l|l|l|}
		\hline
		Grid Label  & Location  & Nominal \\ 
		&&Frequency (Hz) \\ \hline
		A & Texas & 60 \\ \hline
		B & Lebanon & 50 \\ \hline
		C & Eastern U.S. & 60 \\ \hline
		D & Turkey & 50 \\ \hline
		E & Ireland & 50 \\ \hline
		F & France & 50 \\ \hline
		G & Tenerife & 50 \\ \hline
		H & India (Agra)& 50 \\ \hline
		I & Western U.S. & 60 \\ \hline
		J & Brazil& 60 \\ \hline
		K & Norway & 50 \\ \hline
		L & Australia & 50 \\ \hline
	\end{tabular}
	\end{table*}
For each grid, 2 audio recordings of duration 30 min each were provided. The power data were provided for varying amounts of time (5 to 7 hours) for different grids. However, in the development and test data set, both audio and power signals were of 10 minutes duration. Development data consisted of 45 recordings and testing was done on 134 recordings from the 12 grids. The given power data are sinusoidal signals with clear periodicity. However, the audio recordings are extremely noisy, and thus obtaining the time varying power frequency component from such a signal is very challenging. The problem is to build an efficient and accurate classifier system that can detect the region-of-recording for both power and audio recordings. 

\section{Description of the classification system}
A simplified block diagram of the proposed classifier is presented in Fig. \ref{fig:layout}. In what follows we describe 
different blocks sequentially.

\subsection{ENF Extraction Method}
First of all, we determine the nominal frequency and data type (audio/power) of the given test sample. This is done to enhance the inter-grid separability, and thus improve the grid classification performance. 
\begin{figure*}[hh]
	\centering
	\subfloat[STFT magnitude spectrum of a power signal]{\includegraphics[width=.35\textwidth,height=1.7in]{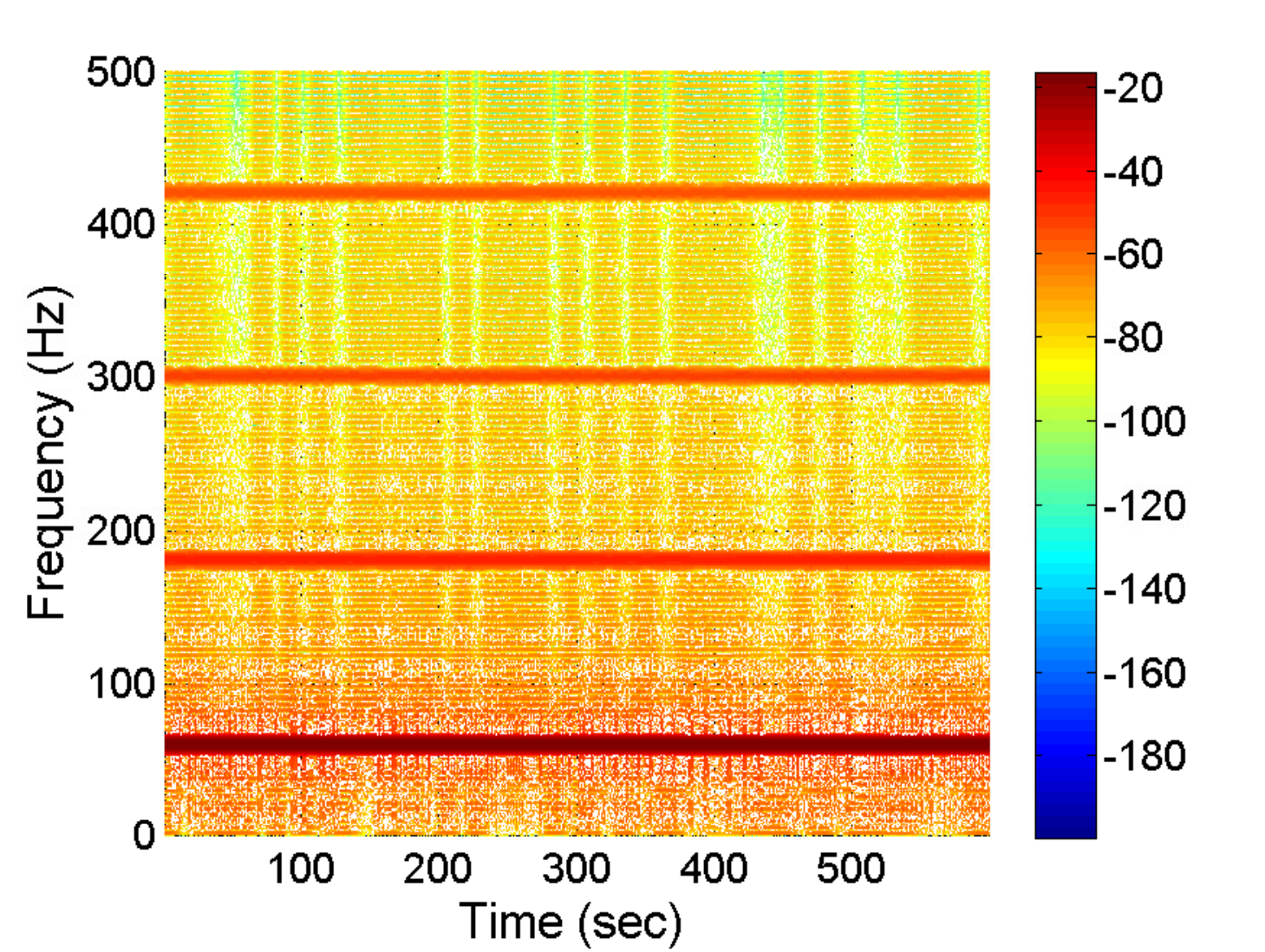}\label{STFT_power}}
	\hfill
	\subfloat[STFT magnitude spectrum of an audio signal]{\includegraphics[width=.35\textwidth,height=1.7in]{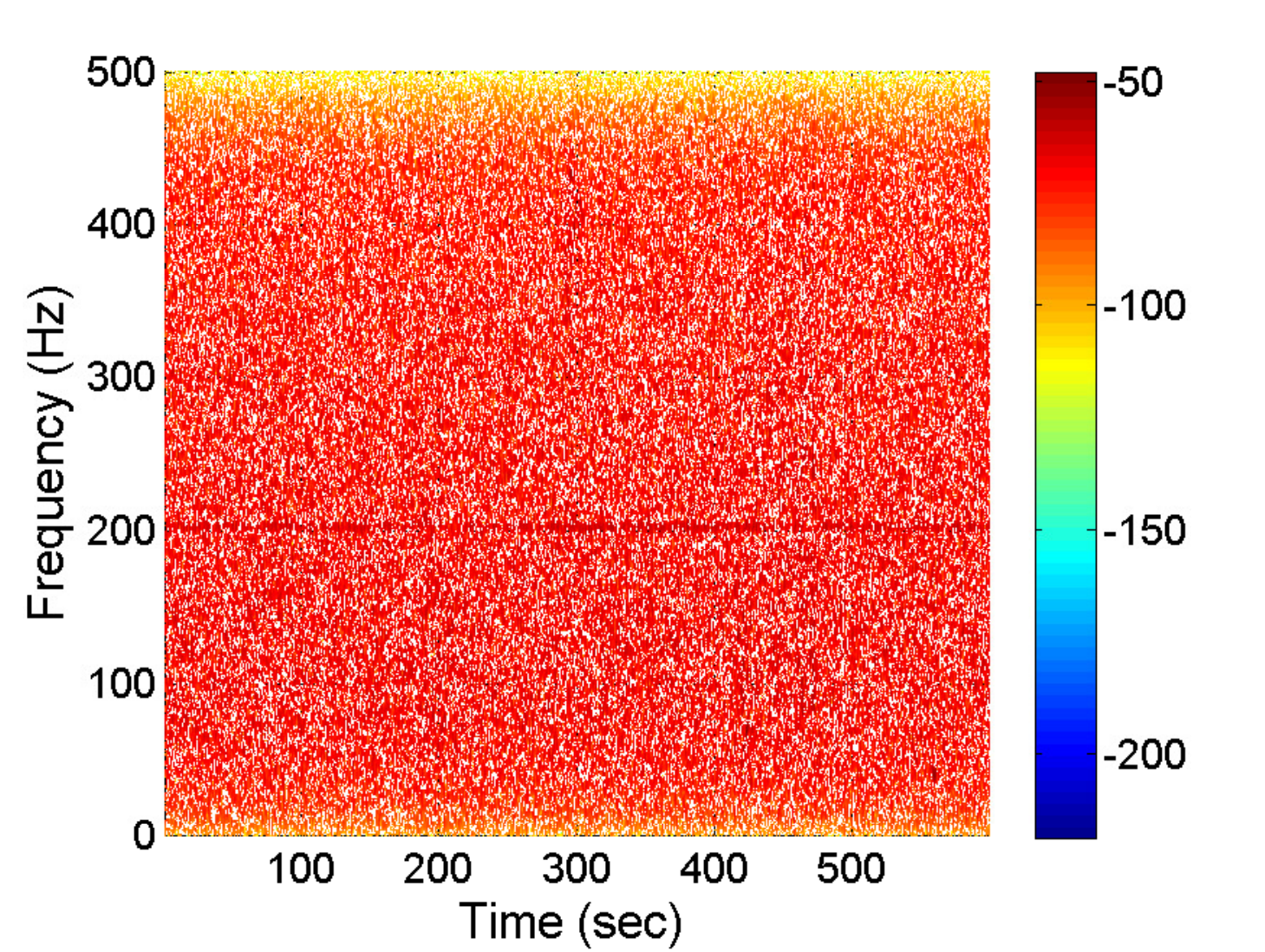}\label{STFT_audio}}
	\caption{Spectrogram estimation using short-time Fourier transform (STFT) of media recordings.}
\end{figure*}
The data type of a given signal can be easily detected in the frequency domain. Figs.~\ref{STFT_power} and \ref{STFT_audio} present the spectrogram of a power and an audio signal, respectively. As we can see, for the power signal, the spectrum is sparse and energy of the signal is mostly concentrated on the nominal and harmonic frequencies. However, for the audio signal, the energy of the signal is spread over the entire spectrum. The figures also reveal that the detection of nominal frequency is a trivial task for the power signal. However, for the audio signal, the ENF signal is covered with the audio signal and hence nominal frequency estimation becomes a challenge. 

In order to determine the nominal frequency, we take the magnitude of the Fourier transform (FT) of the given signal. We split the spectrum into two parts: $S_n$ and $S_r$. The $S_n$ contains the frequency components that are located in the narrow spectral band around the two nominal frequencies and their second harmonics and $S_r$ is the rest of the spectrum. Then we find the frequency of the maximum magnitude in $S_n$ that is denoted by $F_p$. We now calculate two distance parameters for two different nominal frequencies as
\begin{equation}
D_{50}=\min(|F_p-50|, |F_p-100|) 
\end{equation}
and
\begin{equation}
D_{60}=\min(|F_p-60|, |F_p-120|). 
\end{equation}
If $D_{50}<D_{60}$, then 50 Hz is chosen as nominal, otherwise 60 Hz is chosen. 

Again, for audio or power type detection, we compute the summation of frequency components of $S_n$ and $S_r$ that are denoted by $P_n$ and $P_r$, respectively . To calculate $P_r$, we only consider the frequency components up to $125$ Hz. Now, if the ratio of $P_{r}$ to $P_{n}$ is greater than a certain predefined threshold, $T$, the signal is classified as an audio signal. On the other hand, if this ratio is small meaning that the power is highly concentrated on the nominal frequency and its harmonics, then it is detected as a power signal.  

After deciding whether the signal is audio or power, we use different versions of ENF extraction method for each type. For audio signal, we use spectrum combining technique (similar to \citep{hajj2013spectrum} with some modifications). The method in \citep{hajj2013spectrum} includes harmonic analysis which is particularly useful for extracting weak ENF signal from audio data. However, for power data we extract the ENF signal directly from the fundamental nominal frequency band as in \citep{smith1987parshl}. Since the power signal is clean and the nominal frequency signal is always present in the data, we do not need to consider the harmonic bands in this case.

\subsubsection{ENF Extraction from Audio Recording}
For ENF extraction, the total data points (of each recording) are divided into a number of overlapping frames. In our work, the time-duration of each frame is 5s and the overlap between successive frames is 3s. Thus we get estimated ENF values at 2s intervals for the audio signal. In order to obtain the combined spectrum for each frame, we compress and shift the spectral components at different harmonic bands to the nominal base range, $[f_0- f_B, f_0+ f_B]$, and then combine them together as follows (as in \citep{hajj2013spectrum}).
\begin{equation}
S(f) = \sum_{k=1}^{L}w_kP_{B,k}(kf)
\end{equation}
where, $w_k$ is the combining weight of the $k$th harmonic band and $P_{B,k}(f)$ is the spectral band around the $k$th harmonic frequency. As in \citep{hajj2013spectrum}, $w_k$ is obtained by estimating the SNR in the kth harmonic band. The ENF can vary over a wide range across the globe (for example, Laos has $f_0\pm f_B$ = $50\pm 8$ Hz). Therefore, we have chosen three different bandwidths, $f_B = 1, 3, 8$ for SNR calculation. For each value of $f_B$, we get a different $w_k$ and a different combined spectrum $S(f)$. Thus we get three different probable estimates of ENF by searching the maximum in each $S(f)$ for a given time-frame. Finally, we choose the ENF signal that is the least varying among the successive frames. After estimating instantaneous ENF, we perform IIR Hampel filtering followed by smoothing to remove outliers in ENF signals.
\subsubsection{ENF Extraction from Power Recording}
For each recording, total data points are divided into a number of non-overlapping frames of duration $2$s. Then we use an STFT based quadratic interpolation technique to determine the ENF. Choosing the frequency directly from the STFT may give wrong estimation of ENF because Fast Fourier Transform (FFT) is computed for a finite number of discrete frequencies. To overcome this limitation, a quadratic interpolation scheme described in \citep{smith1987parshl} is used in our work.
For each STFT frame, log power spectrum is calculated at discrete frequencies. Then the peak of the spectrum is identified in the frequency range between $46$ Hz to $64$ Hz. Let, $k$ represents the sample index of the peak-spectrum and $\alpha$, $\beta$, $\gamma$ denote the spectrum values at $(k-1)$, $k$ and $(k+1)$. Now according to quadratic interpolation technique, the location of true peak can be estimated as
\begin{equation}
k_{true} = k+p
\end{equation}
where
\begin{equation}\nonumber
p = \frac{\alpha-\gamma}{2(\alpha-2\beta+\gamma)}
\end{equation}
The ENF in Hz is obtained by $(F_s *k_{true})/N$ , where $F_s$ is sampling frequency and $N$ is the number of FFT point used.

\subsection{Feature Ectraction}
The extracted ENF signal is divided into non-overlapping segments having 32 samples in each block from which $38$ features were extracted as shown in Table \ref{feature}.

\begin{table*}[h]
	\centering
	\caption{List of extracted features from the ENF signals.}\label{feature}
	\begin{tabular}{|l|l|}
		\hline
		\textbf{Name of the Feature}  & \textbf{Feature index}\\ \hline \hline
		Variance & 1 \\ \hline
		Mean & 2 \\ \hline
		Average of max. three frequency fluctuation & 3 \\ \hline
		AR Coefficient & 4-5 \\ \hline
		Wavelet Coefficients (5 level and 'Haar' basis) & 6-37 \\ \hline
		Range of each segment & 38 \\ \hline
		\end{tabular}
	\end{table*}

\subsection{SVM Classifier}
We have used four different SVM classifiers for each of the different data types (50Hz-power / 60Hz-power / 50Hz-audio /     60Hz-audio). Different classifiers for different data-types help us to reduce the overlap between different grid features. Moreover, to make the classifier robust against the “curse of dimensionality”, the features are analyzed using the “sparse logistic regression with Bayesian regularization” \citep{cawley2006gene}, and only the significant features are taken. The tolerance level of feature selection was adjusted to $0.6$ by observing the highest cross validation result accuracy. The list of significant features for different data type is presented in Table \ref{featureSelection}. The order of the features implies the level of significance. The most significant feature is placed in the first position. Now, with the selected features for the grids with a specific data type a multiclass SVM with RBF kernel is trained. The hyperparameters for the RBF kernel has been chosen using exhaustive grid search and observing the cross validation accuracy.

\begin{table*}[h]
	\centering
	\caption{List of selected features for different data types.}\label{featureSelection}
	\begin{tabular}{|l|l|}
		\hline
		\textbf{Data Type}  & \textbf{Selected Feature indices.}\\ \hline \hline
		60 Hz power & 1, 38, 5, 3, 13, 4, 20, 26, 22, 34, 30, 6, 37, 27, 19,\\
		            & 31, 36, 32, 33, 18, 14\\ \hline
		50 Hz power & 6, 3, 1, 2, 5, 26, 13, 33, 18, 4, 12, 11, 38, 24, 31, 32,\\
		            & 23, 34, 22, 36, 21, 30, 25, 28, 20, 29\\ \hline
		60 Hz audio & 3, 1, 4, 25, 12, 33, 30, 28, 37, 21 \\ \hline
		50 Hz audio & 2, 1, 3, 26, 37, 4, 6, 11, 22, 13, 28, 29, 5, 35, 10,\\
		            & 31\\ \hline
		\end{tabular}
	\end{table*}
	
In what follows, we describe how our SVM classifier works. We have $9$ segments of ENF with each having $32$ samples as described in the previous section for a test sample. First, an SVM multiclass classifier trained for the corresponding data type gives a probability estimate for each segment of ENF. Some of these segments will highly deviate from the feature space of the outlier (incorrect) grids leading to a poor probability estimate (close to $0$), whereas all the segments will give high probability estimates for the probable correct grids. Therefore, we use the geometric mean of the probability estimates of the $9$ segments as the overall probability estimate so as to reduce the overall probability estimate for the outlier grids to a greater extent. Now, using the probability estimates of the SVM for the test sample, we narrow down the list of two ($60$ Hz) or three ($50$ Hz) most probable grids, and the decision is then passed to the pole matching classifier to take final decision.

\subsection{Pole Estimation}
So far, we have been concerned about the instantaneous nominal frequency of a grid, i.e., ENF. However, interaction between a power grid and the loads under it, have an impact on the fluctuation of the grid frequency as well as on the voltage magnitude leading to different transient phenomena like surge, sag, swell \citep{gruzspower} in the power grid. These fluctuation patterns of voltage magnitude are unique to a particular grid as it depends on the loading schedule specific to that grid. In addition to the voltage magnitude fluctuation pattern, different grids have different nominal values of voltage at the power outlet and distribution lines in different parts of the world such as most Asian grids have a nominal voltage magnitude of $220$ volts whereas USA maintains a nominal voltage of $120$ volts at its power outlets. Therefore, in addition to ENF pattern, voltage fluctuation pattern and its nominal value can be very promising in increasing the accuracy of an origin-of-recording identification system. Considering this in what follows, we attempt to use parametric modeling, i.e., AR modeling to model the power grid as a linear system driven by white noise. An AR model is suitable for signals that have sudden peaks in their frequency spectrum \citep{guler2001ar} similar to power signal which has peaks at the nominal frequency along with its harmonics. Again, the roots of the AR parameters reveal the positions of the power grid poles in the z-plane that reflect both the frquency and the voltage magnitude of the signal. To estimate the pole locations of the data, we model the given raw data, $x(n)$, as the output of an autoregressive (AR) system as 
\begin{equation}
x(n)=\sum_{k=1}^{N}a_k x(n-k) + e(n)
\label{AR_model}
\end{equation}
where, $N$ is the order of the AR system, $a_k$ represent the AR coefficients and e(n) is the random excitation signal. Applying z-transform on both side of (\ref{AR_model}), we can get the transfer function, $H(z)$, of an all-pole AR system as 
\begin{equation}
H(z)=\frac{X(z)}{E(z)}=\frac{1}{1-\sum_{k=1}^{N}a_k z^{-k}} 
\label{AR_tf}
\end{equation}
The roots of the denominator polynomial of (\ref{AR_tf}) provides us the poles of the given data $x(n)$. However, we need to evaluate the values of AR coefficients $a_k$ to estimate the poles. For this, we multiply both sides of (\ref{AR_model}) by $x(n-j)$ and take the expectation operation which gives
\begin{equation}
E[x(n)x(n-j)]=\sum_{k=1}^{N}a_k E[x(n-k)x(n-j)] + E[e(n)x(n-j)]
\label{AR_expect}
\end{equation}
where $E[.]$ represent expectation operation. Now for the optimal model coefficients, the random excitation $e(n)$ is orthogonal to the past samples, and (\ref{AR_expect}) gives
\begin{equation}
r_{xx}(j)=\sum_{k=1}^{N}a_k r_{xx}(j-k). 
\label{AR_corr}
\end{equation}
Given $N+1$ correlation values, (\ref{AR_corr}) can be solved to obtain the AR coefficients $a_k$. Since the data is non-stationary, i.e., the grid frequency varies with time, it is not useful to compute the correlation values for the whole data. Therefore, we divide the given data into non-overlapping small segments assuming that the grid frequency remains stationary during that short period. Now, for each segment, the correlation values are estimated that give the AR coefficients, and finally, we get the estimated poles. The poles extracted from the training dataset are stored as reference to be compared with those obtained from the test data for grid identification.

\subsection{Pole-matching Classifier}
\begin{figure*}[t]	
	\centering
	\includegraphics[width=3.0in]{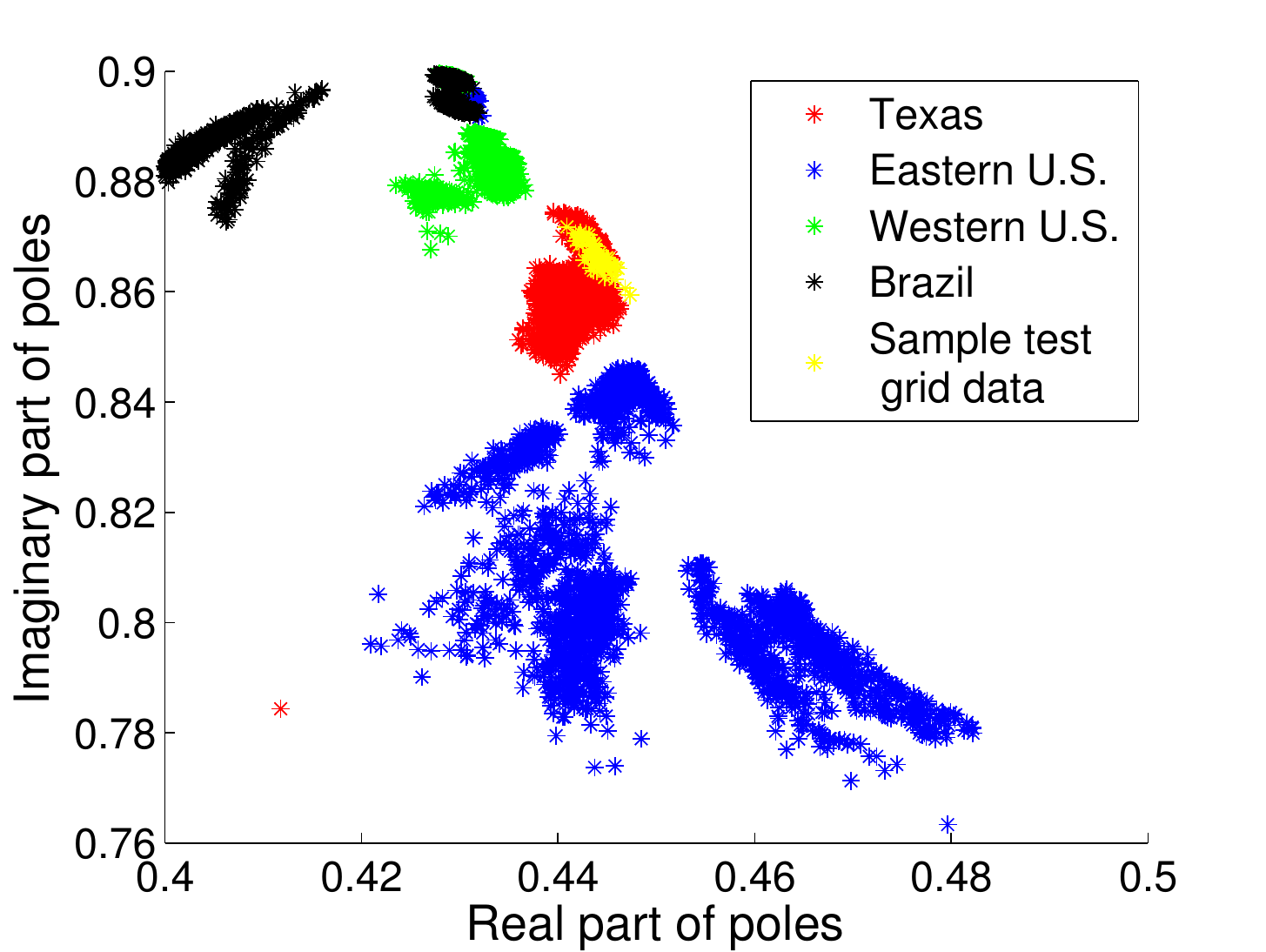}\\
	\caption{Plot of the location of poles of a testing set superimposed on the pole plot of the power recordings of the training grids.}\label{pole_match_ex}
\end{figure*}
First we explain the concept of our pole-matching classifier through an illustrative example as shown in Fig.~\ref{pole_match_ex}. Here, the pole locations of a test signal is plotted along with four different training poles. The different training poles are marked with different colors and they are extracted from Texas, Eastern U.S., Western U.S., and Brazil grids. It is apparent that the test poles (in yellow) match mostly with the poles of the Texas grid (marked as red). Therefore, we can easily decide that this test signal belongs to the Texas grid. The proposed pole-matching classifier is based on this visual comprehension.

The test signal is first segmented into short duration of $10$ seconds with non-overlapping blocks, and the poles are estimated from each blocks. After that, the classification process is initiated by measuring distances between the pole points of test signal and those of the training dataset. Let, $p_{i}$ is the $i^{th}$ pole point of the test data and $\mathbf{g}^k$ is the vector containing all the pole points of the training data of grid $k$. Then, a distance function is calculated as follows
\begin{equation}
d_{i,j}^{k} = ||p_{i} - {g}_j^k||,
\end{equation}
where $||.||$ denote the $l_2$ norm, $i=1,2,....U$, $j=1,2,....V_k$ and $k=1,2,....R$. Here, $U,~V_k$ and~$R$ are total no.~of testing poles, total no.~of training poles of grid $k$ and total no.~of training grids, respectively. For each testing pole, only $X$ no.~of minimum distances are considered and stored in a vector. Thus, a vector $\mathbf{d}^k$ of length $XU$ is obtained for the grid k considering all the test poles. We then calculate the average distance ${o}^{k}$ from $\mathbf{d}^k$ as
\begin{equation}
o^{k} = \frac{1}{XU}\sum_{l=1}^{XU} d^k(l).
\end{equation}
The final grid label for the test data is determined from the grid that gives the minimum distance. Mathematically, the identified grid is
\begin{equation}
\widehat{grid} = \underset{k}{\arg\min}(o^{k}).
\end{equation}

\begin{table*}[h]
	\centering
	\caption{Meta data of four test samples used to verify the grid detection results.}\label{meta_data}
	\begin{tabular}{|l|l|l|l|}
		\hline
		Sample  & Data & Nominal   & Region of   \\ 
		No.     & type & frequency & recording  \\ 
		\hline
		
		1 & Power & 60 & Eastern US\\ \hline
		2 & Audio & 60 & Texus\\ \hline
		3 & Power & 50 & Ireland \\ \hline
		4 & Audio & 50 & Agra-India \\ \hline
	\end{tabular}
\end{table*} 

\begin{table*}[h]
	\centering
	\caption{Numerical results of detection of data type and nominal frequency for the test samples.}\label{TypenNominal}
	\begin{tabular}{|l|l|l|l|l|l|}
		\hline
		Sample  & \multicolumn{2}{|l|}{Data type } & \multicolumn{3}{|l|}{Nominal frequency }  \\ \cline{2-6} 
		No. & $P_{r}/P_{n}$   & Data  & $D_{50}$ & $D_{60}$ & Nominal   \\
		    &                                & type   &              &               & freq.(Hz)\\ \hline 
		
		1 & 0.375 & Power & 10.005  & 0.035 & 60 \\ \hline
		2 & 7.620 & Audio  &  19.965 &  0.035 & 60\\ \hline
		3 & 0.390 & Power & 0.052 & 10.052 & 50 \\ \hline
		4 & 10.741 & Audio & 0.323 & 19.677 & 50\\ \hline		
		
\end{tabular}
\end{table*} 

\begin{figure*}[ht!]
\centering
\includegraphics[width = 5.5 in, height = 4.1 in]{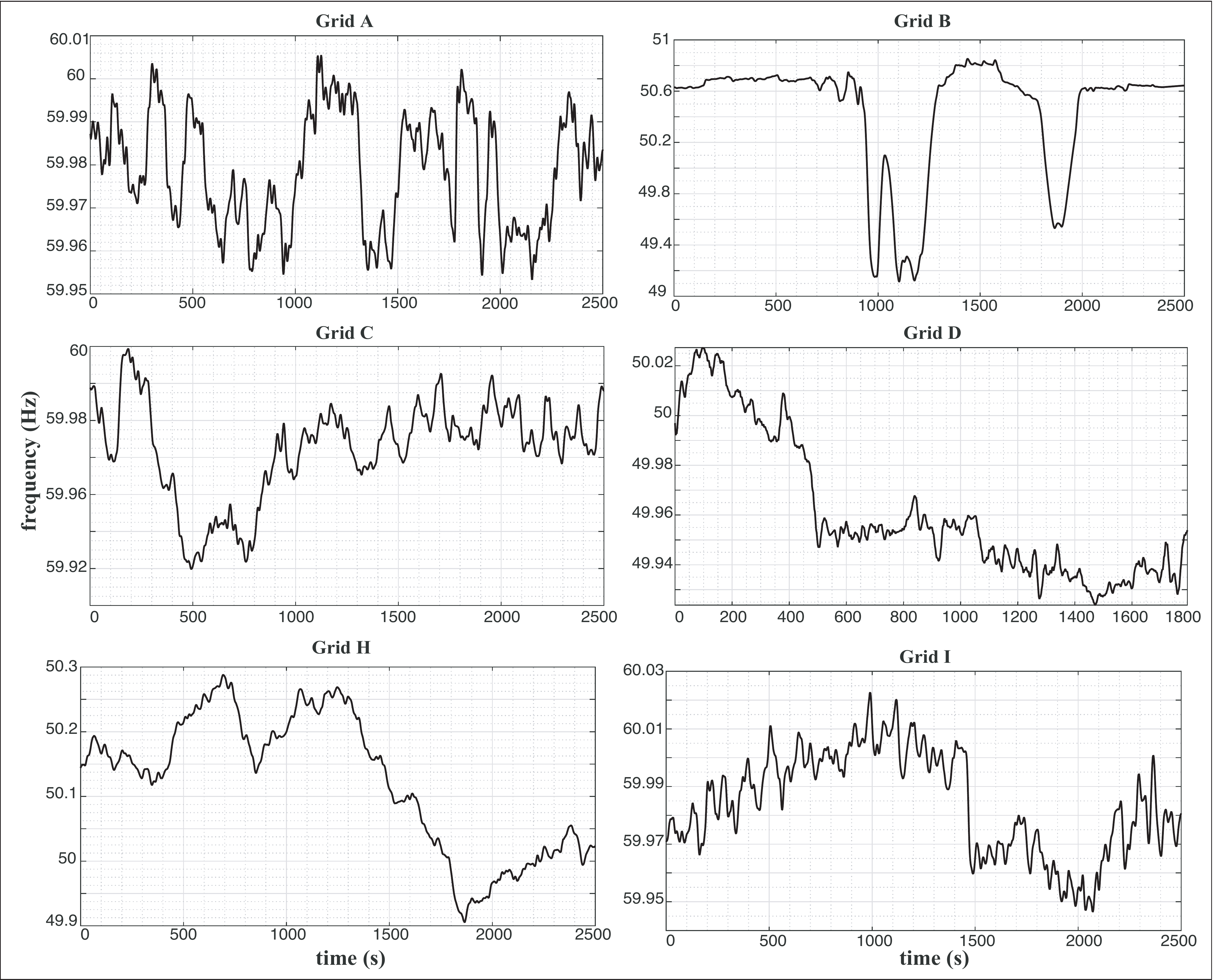}
\caption{ENF signals of different grids extracted from the power data.}
\label{fig:enf_pow}
\end{figure*}

\begin{figure*}[ht!]
\centering
\includegraphics[width = 5.5 in, height = 4.1 in]{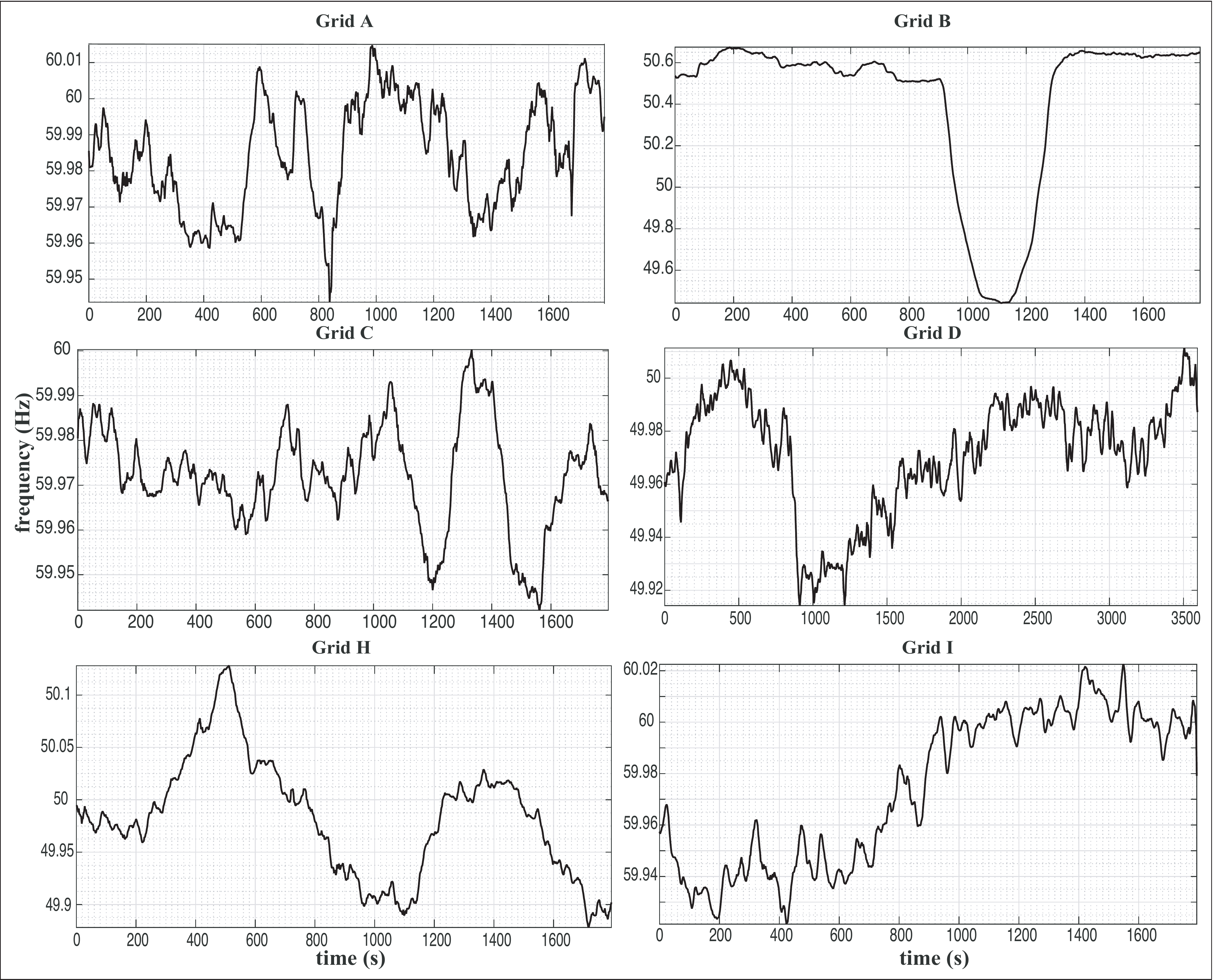}
\caption{ENF signals of different grids extracted from the audio data.}
\label{fig:enf_audio}
\end{figure*}

\begin{figure*}[ht!]
\centering
\includegraphics[width = 3.5 in, height = 5.2 in]{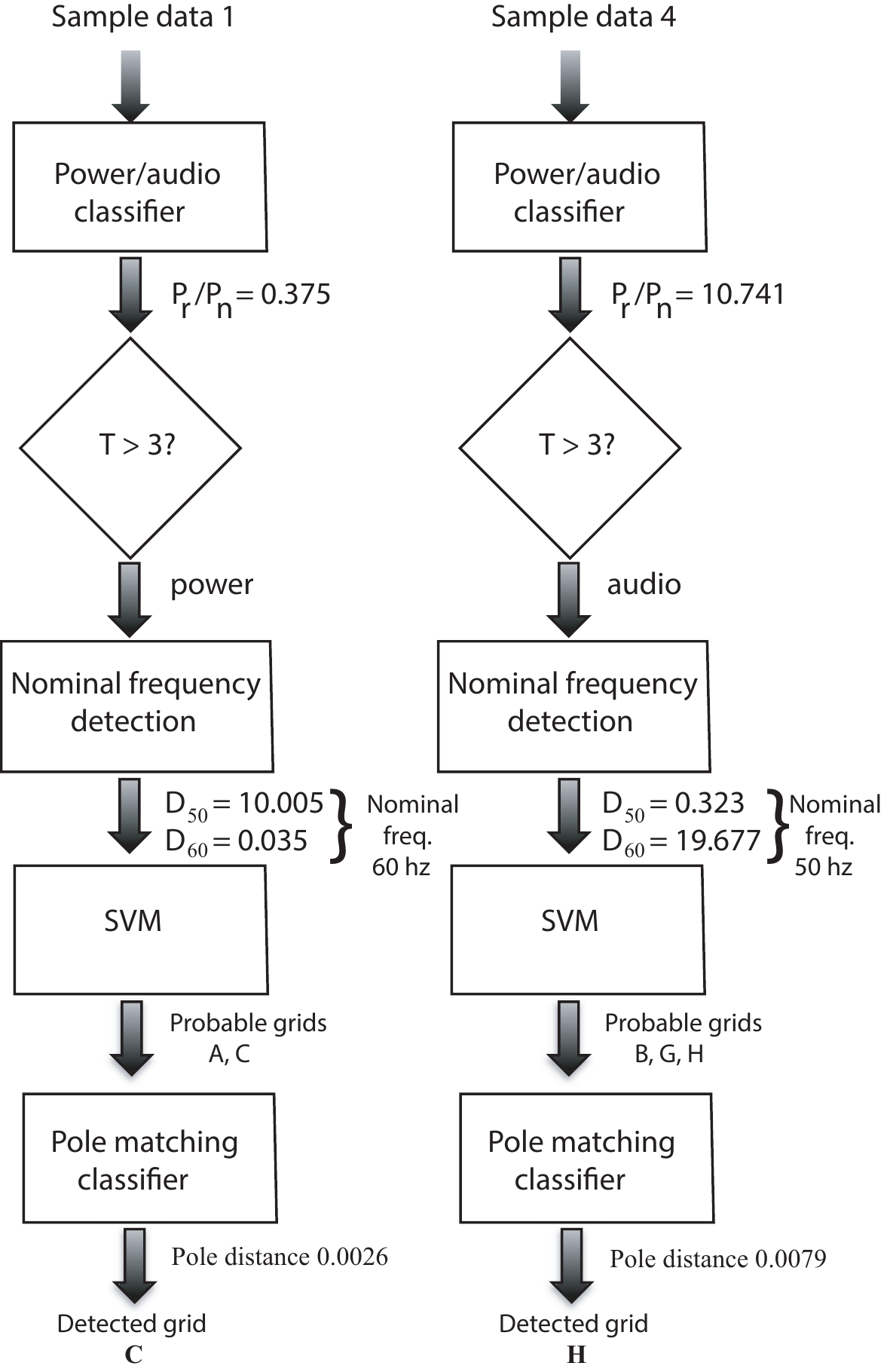}
\caption{Workflow diagram of two sample data gird of origin detection.}
\label{fig:workflow}
\end{figure*}

\begin{table*}[h]
	\centering
	\caption{Pole matching distance between test poles and available training grids.}\label{sample_dist_measure}
	\begin{tabular}{|l|l|l|l|l|l|l|l|}
		\hline
		Test sample  & \multicolumn{6}{|l|}{Available grids} & \multirow{2}{*}{Detected grid}  \\ \cline{1-7} 
	60 Hz  & A  & C  & I  &     &    &    &      \\ \hline 
		1 & 0.0373 & 0.0026   & \_ &  &    &    &    C (Eastern US) \\ \hline
		2 & 0.0024 & \_   & 0.0124 &  &    &    &    A (Texus)\\ \hline
        50 Hz  & B & D & E & F & G & H &    \\ \hline
		3 & 0.0036 & 0.0014   & 9.13e-5 & \_~~~~~ & \_ & \_ &  E (Ireland) \\ \hline
		4 & 0.0817 & \_   & \_ & \_ & 0.0882 & 0.0079 &  H (Agra-India)\\ \hline		
\end{tabular}
\end{table*} 

\section{Results}
In this section we present results to validate the effectiveness of our grid identification system and performance accuracy analysis.
\subsection{Examples of Extracted ENF Signals}
Fig. \ref{fig:enf_pow} and \ref{fig:enf_audio} show some sample ENFs extracted from the power and the audio data using our adopted methods for different $60$ Hz and $50$ Hz grids. It is observed that the ENF of a particular grid is significantly different from the other in terms of variance, mean, maximum rate of change, and fluctuation range, and thus the ENF signal qualifies to be the power signature of a particular grid. Again, though the nominal frequency of A, C, and I grids are same, the mean and variance of the ENF will be different. The same observation is valid for B, D, and H grids.

\subsection{Parameter Settings}
First, we mention the specified values of different parameters used in the experiment. The threshold, $T$, for power or audio data type detection was set to $3$. The data was segmented into 10 sec blocks for pole estimations. The order of the AR system was 8 and 12 for power and audio data, respectively. The pole matching distance for a certain grid was calculated using $X=2$, i.e., the mean distance of the two closest training poles from each test pole was used for measuring pole-matching distance.  

\subsection{Performance Measurement}
The performance of different grid detection algorithms are evaluated using the accuracy criterion defined as
\begin{equation}
Accuracy (\%)=\frac{No.~of~correctly~classified~grids}{Total~no.~of~grids ~classified}
\end{equation}

\begin{table*}[ht!]
	\centering
	\caption{Distance of different segments of the test ENF from the centroid of training data of F and G grids}\label{distance}
	\begin{tabular}{|l|l|l|}
		\hline
		Test ENF & \multicolumn{2}{|l|}{Distance from grid}\\ \cline{2-3} 
		segment no. & F~~~~~~~ & G \\ \hline 
		1 & 2.1108 & 1.9566\\ \hline
		2 & 1.4139 & 1.3583\\ \hline
		3 & 1.9378 & 1.7382\\ \hline
		4 & 0.2413 & 0.2041\\ \hline
		5 & 0.0896 & 0.3428\\ \hline
		6 & 1.2921 & 1.0468\\ \hline
		7 & 0.9381 & 0.7430\\ \hline
		8 & 0.1370 & 0.3645\\ \hline
		9 & 1.0622 & 0.7796\\ \hline
	\end{tabular}
\end{table*} 

\begin{table*}[ht!]
	\centering
	\caption{Classification accuracy on different grids using pole-matching, ENF-based SVM and cascaded classifier for the testing set. Here the digit in the parenthesis denote the total no. of files of that type and the digit before the parenthesis denote the no. of correctly identified datafile}\label{Tabaccuracy}
	\begin{tabular}{|l|l|l|l|l|l|l|}
		\hline
		Grid  & \multicolumn{3}{|l|}{ENF-based SVM} & \multicolumn{3}{|l|}{Proposed cascaded classifier} \\ \cline{2-7} 
		label & Power  & Audio & Accuracy($\%$) & Power & Audio & Accuracy($\%$) \\ 
		      & $\#$ correct & $\#$ correct &  & $\#$ correct  & $\#$ correct & $\#$ correct\\
		      & ($\#$ total) & ($\#$ total) &  & ($\#$ total)  &  ($\#$ total) & ($\#$ total)\\ \hline 
		
		A & 9(9) & 2(3) & 92 & 9(9) & 3(3) & 100\\ \hline
		B & 8(8) & 6(6) & 100 & 8(8) & 6(6) & 100\\ \hline
		C & 5(8) & 3(7) & 53 & 7(8) & 7(7) & 94\\ \hline
		D & 6(7) & 7(7) & 93 & 6(7) & 7(7) & 93\\ \hline
		E & 7(8) & 2(5) & 69 & 7(8) & 3(5) & 77\\ \hline
		F & 6(8) & 3(3) & 82 & 8(8) & 3(3) & 100\\ \hline
		G & 6(9) & 2(6) & 53 & 9(9) & 6(6) & 100\\ \hline
		H & 8(8) & 4(5) & 92 & 8(8) & 5(5) & 100\\ \hline
		I & 6(8) & 7(7) & 87 & 8(8) & 6(7) & 94\\ \hline	
		All & 61(73) & 36(49) & 79.51 & 70(73) & 46(49) & 95.08\\ \hline
		
	\end{tabular}
\end{table*} 

\subsection{Verification of Grid Identification Results}
Here, we describe the workflow of our proposed grid identification algorithm using numerical examples. We consider four test samples representing power and audio data type and 50 or 60 Hz nominal frequencies. The meta data associated with the test samples are presented in Table~\ref{meta_data}. Different parameters extracted from the test samples for data type and nominal frequency classification are shown in Table~\ref{TypenNominal}. It is seen that data type and nominal frequency can be easily determined from the extracted parameters. Next, the number of probable grids is narrowed down using the probability estimates obtained from SVM. Table~\ref{sample_dist_measure} shows the list of probable grids for each test samples along with the calculated pole matching distances. The pole-matching classifier determines the final estimated grid based on the minimum distance. A complete workflow of the overall procedure for the detection of sample data $1$ and $4$ from Table~\ref{meta_data} is shown in Fig. \ref{fig:workflow}.

\subsection{Advantage of Cascaded Classifiers}
In this section, we attempt to explain the intuition behind our proposed cascaded classification system. For the detection purpose, the ENF extracted from a test sample is split into short duration segments that may be similar to more than one grid in terms of features. To validate our claim, we choose a $50$ Hz power metadata of grid F from the test dataset and extract the feature vectors for all the $9$ segments of the ENF signal. As visualization of higher dimensional data is difficult, we calculate the centroid, i.e., mean position of a multivariate distribution, of the feature vectors extracted from the training data of power grid F and G. Then we calculate the Euclidean distance between the feature vectors of the test data and the centroid of F and G grids as a measure of similarity between the test and train data. It is seen from Table \ref{distance} that the test sample is close to both F and G grids in terms of the considered feature sets extracted from the ENF, and unanimous decision in favour of any grid is difficult to take. From the above discussion, it is clear that the metadata has ENF pattern overlapping in terms features with both grid F and G, and hence, SVM gives mean probability estimates of $0.631$ and $0.369$ for G and F grids respectively whereas the metadata was from grid F. Therefore, only ENF based classification may fail to classify a grid successfully if it has an ENF pattern similar to more than one grid. In our proposed scheme, we propose to use the SVM classifier as the front-end and narrow down the probable grids using it. Thereafter, we use the PM classifier which incorporates both the frequency and the voltage magnitude fluctuation pattern of a grid to detect the correct grid. PM classifier gives pole distance of $0.0001844$ and $0.0048$ for grid F and G, respectively and thereby, successfully detects the test sample as of grid F.

\subsection{Comparison with ENF-based and Proposed Cascaded Classifiers}
Table \ref{Tabaccuracy} presents the grid identification results of cascaded classifiers (proposed ) and ENF-based SVM classifier (baseline) on the given test set. The baseline system has $83.56$\% and $73.47$\% accuracy on the power and audio samples, respectively. To the contrary, the proposed system achieves $95.89$\% and $93.88$\% accuracy on the power and audio samples, respectively. On the average, the proposed cascaded classifiers give superior accuracy of $95.08$\% compared to $79.51$\% of only ENF-based SVM classifier.

\section{Conclusion}
In this paper, we have developed a classification system based on cascaded SVM and pole-matching criterion that can identify the grid-of-origin of a power or audio signal. The proposed approach makes use of the estimated poles calculated from the raw power or audio data that are used in a distance based classification system with a front-end of conventional ENF-based SVM classifiers. This proposed cascaded classification system shows much higher accuracy as discussed in the result section compared to that of standalone ENF-based SVM classifier. 




\newpage
 \section*{References}
\bibliographystyle{elsarticle-num-names} 
\bibliography{jd}

\begin{thebibliography}{18}
\expandafter\ifx\csname natexlab\endcsname\relax\def\natexlab#1{#1}\fi
\providecommand{\url}[1]{\texttt{#1}}
\providecommand{\href}[2]{#2}
\providecommand{\path}[1]{#1}
\providecommand{\DOIprefix}{doi:}
\providecommand{\ArXivprefix}{arXiv:}
\providecommand{\URLprefix}{URL: }
\providecommand{\Pubmedprefix}{pmid:}
\providecommand{\doi}[1]{\href{http://dx.doi.org/#1}{\path{#1}}}
\providecommand{\Pubmed}[1]{\href{pmid:#1}{\path{#1}}}
\providecommand{\bibinfo}[2]{#2}
\ifx\xfnm\relax \def\xfnm[#1]{\unskip,\space#1}\fi
\bibitem[{UNICEF et~al.(2011)}]{unicef2011child}
\bibinfo{author}{UNICEF}, et~al.,
\newblock \bibinfo{title}{Child protection from violence, exploitation and
  abuse child trafficking},
\newblock \bibinfo{journal}{URL: http://www. unicef. org/protection/57929}
  \bibinfo{volume}{58005} (\bibinfo{year}{2011}).
\bibitem[{Garg et~al.(2013)Garg, Hajj-Ahmad, and Wu}]{garg2013geo}
\bibinfo{author}{R.~Garg}, \bibinfo{author}{A.~Hajj-Ahmad},
  \bibinfo{author}{M.~Wu},
\newblock \bibinfo{title}{Geo-location estimation from electrical network
  frequency signals.},
\newblock in: \bibinfo{booktitle}{ICASSP}, \bibinfo{year}{2013}, pp.
  \bibinfo{pages}{2862--2866}.
\bibitem[{Fechner and Kirchner(2014)}]{fechner2014humming}
\bibinfo{author}{N.~Fechner}, \bibinfo{author}{M.~Kirchner},
\newblock \bibinfo{title}{The humming hum: Background noise as a carrier of
  {ENF} artifacts in mobile device audio recordings},
\newblock in: \bibinfo{booktitle}{IT Security Incident Management \& IT
  Forensics (IMF), 2014 Eighth International Conference on},
  \bibinfo{organization}{IEEE}, \bibinfo{year}{2014}, pp.
  \bibinfo{pages}{3--13}.
\bibitem[{Grigoras(2005)}]{grigoras2005digital}
\bibinfo{author}{C.~Grigoras},
\newblock \bibinfo{title}{Digital audio recording analysis--the electric
  network frequency criterion},
\newblock \bibinfo{journal}{International Journal of Speech Language and the
  Law} \bibinfo{volume}{12} (\bibinfo{year}{2005}) \bibinfo{pages}{63--76}.
\bibitem[{Cooper(2008)}]{cooper2008electric}
\bibinfo{author}{A.~J. Cooper},
\newblock \bibinfo{title}{The electric network frequency ({ENF}) as an aid to
  authenticating forensic digital audio recordings--an automated approach},
\newblock in: \bibinfo{booktitle}{Audio Engineering Society Conference: 33rd
  International Conference: Audio Forensics-Theory and Practice},
  \bibinfo{organization}{Audio Engineering Society}, \bibinfo{year}{2008}.
\bibitem[{Grigoras(2007)}]{grigoras2007applications}
\bibinfo{author}{C.~Grigoras},
\newblock \bibinfo{title}{Applications of {ENF} criterion in forensic audio,
  video, computer and telecommunication analysis},
\newblock \bibinfo{journal}{Forensic science international}
  \bibinfo{volume}{167} (\bibinfo{year}{2007}) \bibinfo{pages}{136--145}.
\bibitem[{Cooper(2009)}]{cooper2009automated}
\bibinfo{author}{A.~J. Cooper},
\newblock \bibinfo{title}{An automated approach to the electric network
  frequency ({ENF}) criterion: theory and practice.},
\newblock \bibinfo{journal}{International Journal of Speech, Language \& the
  Law} \bibinfo{volume}{16} (\bibinfo{year}{2009}).
\bibitem[{Grigoras(2009)}]{grigoras2009applications}
\bibinfo{author}{C.~Grigoras},
\newblock \bibinfo{title}{Applications of {ENF} analysis in forensic
  authentication of digital audio and video recordings},
\newblock \bibinfo{journal}{Journal of the Audio Engineering Society}
  \bibinfo{volume}{57} (\bibinfo{year}{2009}) \bibinfo{pages}{643--661}.
\bibitem[{Cooper(2011)}]{cooper2011further}
\bibinfo{author}{A.~J. Cooper},
\newblock \bibinfo{title}{Further considerations for the analysis of {ENF} data
  for forensic audio and video applications.},
\newblock \bibinfo{journal}{International Journal of Speech, Language \& the
  Law} \bibinfo{volume}{18} (\bibinfo{year}{2011}).
\bibitem[{Garg et~al.(2011)Garg, Varna, and Wu}]{garg2011seeing}
\bibinfo{author}{R.~Garg}, \bibinfo{author}{A.~L. Varna},
  \bibinfo{author}{M.~Wu},
\newblock \bibinfo{title}{Seeing {ENF}: natural time stamp for digital video
  via optical sensing and signal processing},
\newblock in: \bibinfo{booktitle}{Proceedings of the 19th ACM international
  conference on Multimedia}, \bibinfo{organization}{ACM}, \bibinfo{year}{2011},
  pp. \bibinfo{pages}{23--32}.
\bibitem[{Liu et~al.(2012)Liu, Yuan, Markham, Conners, and
  Liu}]{liu2012application}
\bibinfo{author}{Y.~Liu}, \bibinfo{author}{Z.~Yuan}, \bibinfo{author}{P.~N.
  Markham}, \bibinfo{author}{R.~W. Conners}, \bibinfo{author}{Y.~Liu},
\newblock \bibinfo{title}{Application of power system frequency for digital
  audio authentication},
\newblock \bibinfo{journal}{IEEE Transactions on Power Delivery}
  \bibinfo{volume}{27} (\bibinfo{year}{2012}) \bibinfo{pages}{1820--1828}.
\bibitem[{Ojowu et~al.(2012)Ojowu, Karlsson, Li, and Liu}]{ojowu2012enf}
\bibinfo{author}{O.~Ojowu}, \bibinfo{author}{J.~Karlsson},
  \bibinfo{author}{J.~Li}, \bibinfo{author}{Y.~Liu},
\newblock \bibinfo{title}{{ENF} extraction from digital recordings using
  adaptive techniques and frequency tracking},
\newblock \bibinfo{journal}{IEEE Transactions on Information Forensics and
  Security} \bibinfo{volume}{7} (\bibinfo{year}{2012})
  \bibinfo{pages}{1330--1338}.
\bibitem[{Hajj-Ahmad et~al.(2013)Hajj-Ahmad, Garg, and Wu}]{hajj2013spectrum}
\bibinfo{author}{A.~Hajj-Ahmad}, \bibinfo{author}{R.~Garg},
  \bibinfo{author}{M.~Wu},
\newblock \bibinfo{title}{Spectrum combining for {ENF} signal estimation},
\newblock \bibinfo{journal}{IEEE Signal Processing Letters}
  \bibinfo{volume}{20} (\bibinfo{year}{2013}) \bibinfo{pages}{885--888}.
\bibitem[{spc(2016)}]{spcup}
\bibinfo{title}{Signal processing cup 2016, database. [online]. available:
  http://sigport.org/documents/information-mast-enf-power-signature- dataset.}
  (\bibinfo{year}{2016}).
\bibitem[{Smith and Serra(1987)}]{smith1987parshl}
\bibinfo{author}{J.~O. Smith}, \bibinfo{author}{X.~Serra},
  \bibinfo{title}{PARSHL: An analysis/synthesis program for non-harmonic sounds
  based on a sinusoidal representation}, \bibinfo{publisher}{CCRMA, Department
  of Music, Stanford University}, \bibinfo{year}{1987}.
\bibitem[{Cawley and Talbot(2006)}]{cawley2006gene}
\bibinfo{author}{G.~C. Cawley}, \bibinfo{author}{N.~L. Talbot},
\newblock \bibinfo{title}{Gene selection in cancer classification using sparse
  logistic regression with bayesian regularization},
\newblock \bibinfo{journal}{Bioinformatics} \bibinfo{volume}{22}
  (\bibinfo{year}{2006}) \bibinfo{pages}{2348--2355}.
\bibitem[{Gruzs(1990)}]{gruzspower}
\bibinfo{author}{T.~M. Gruzs},
\newblock \bibinfo{title}{power qualm}  (\bibinfo{year}{1990}).
\bibitem[{G{\"u}ler et~al.(2001)G{\"u}ler, Kiymik, Akin, and
  Alkan}]{guler2001ar}
\bibinfo{author}{I.~G{\"u}ler}, \bibinfo{author}{M.~K. Kiymik},
  \bibinfo{author}{M.~Akin}, \bibinfo{author}{A.~Alkan},
\newblock \bibinfo{title}{Ar spectral analysis of eeg signals by using maximum
  likelihood estimation},
\newblock \bibinfo{journal}{Computers in biology and medicine}
  \bibinfo{volume}{31} (\bibinfo{year}{2001}) \bibinfo{pages}{441--450}.

\end{thebibliography}


\end{document}